\pdfoutput=1
\documentclass[11pt]{article}

\usepackage[utf8]{inputenc} % From ACL template
\usepackage{times} % From ACL template
\usepackage{latexsym} % From ACL template
\usepackage[dvipsnames]{xcolor} % For color names
\usepackage{amsmath} % Extended math typesetting
\usepackage{amssymb} % Extended math symbols
\usepackage{mathtools} % More mathtools
\usepackage{annotate-equations} % Help annotate equations
\usepackage{pifont} % More symbols (also needed for checkmark and x-mark symbol macros)
\usepackage{truncate} % Help truncate text
\usepackage{enumitem} % allows setting leftmargin on \begin{enumerate} or \begin{itemize}
\usepackage{csquotes} % Inline and display quotes
\usepackage{xurl} % For \url
\usepackage[dont-mess-around]{fnpct} % Automatically ensures footnotes go after the punctuation mark
\usepackage{soul} % For text highlighting
\usepackage{ragged2e} % allows \centering
\usepackage{float} % allows the [H] on \begin{figure}[H]

\usepackage{caption}
\usepackage{subcaption} % For subfigures / captions: https://tex.stackexchange.com/a/122329

\usepackage{graphicx} % allows \includegraphics

\usepackage{array}
\usepackage{longtable}
\usepackage{booktabs}
\usepackage{multirow}
\usepackage{arydshln}
\usepackage{tabularx}
\usepackage{tabulary}
\usepackage{makecell}

\usepackage{tikz,pgfplots}
\pgfplotsset{compat=1.17}

\usepackage{xinttools} % Useful to populate table data with a loop
\usepackage{ifthen} % If-then in LaTeX

\usepackage{lipsum}

\usepackage{todonotes}

\usepackage[T1]{fontenc}
\newcommand{\cmmnt}[1]{}

\newcommand{\mainfig}{
    \begin{figure}[ht!]
        \centering
        \includegraphics[width=185px]{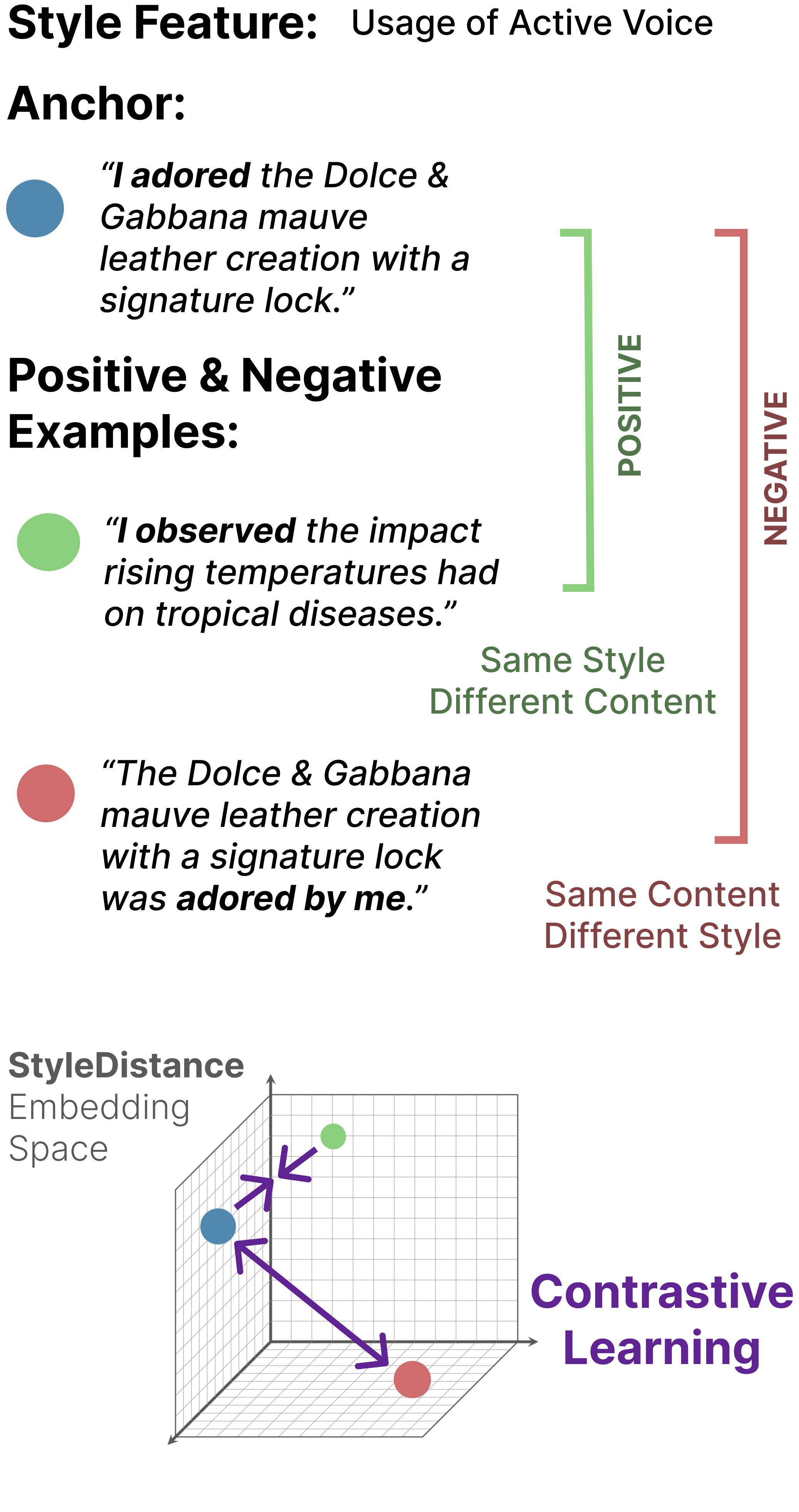}
        \caption{\textsc{StyleDistance} embeddings are trained using contrastive learning from synthetic parallel (positive and negative) examples %synthetically generated by a LLM that are positive and negative examples of usage of a style feature. The synthetic dataset % \textsc{StyleDistance} is trained on 
        representing 40  style features. The illustrated example is for the ``Usage of Active Voice'' feature.}
        \label{fig:main}
    \end{figure}
}

\newcommand{\mturkinterfacefig}{
    \begin{figure}[H]
        \centering
        \includegraphics[width=.9\linewidth]{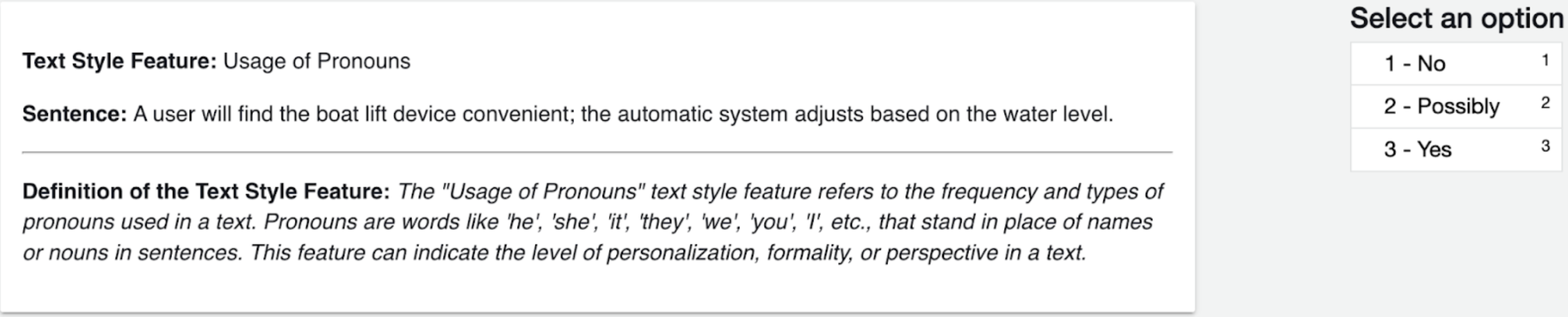}
        \caption{The annotation interface used for human annotation.}
        \label{fig:mturkinterface}
    \end{figure}
}

\newcommand{\stylefeaturecategoriesfig}{
    \begin{figure*}[ht!]
        \centering
        \resizebox{0.95\textwidth}{!}{
            \begin{tikzpicture}[
                font=\sffamily\scriptsize,
                category/.style={rectangle, draw, minimum width=4cm, inner sep=10pt, align=left, rounded corners=3pt}, % Same width within columns
                feature/.style={align=left}
            ]
            
            \definecolor{color1}{RGB}{230,230,250}  % Light lavender
            \definecolor{color2}{RGB}{255,239,213}  % Light peach
            \definecolor{color3}{RGB}{240,255,240}  % Light mint green
            \definecolor{color4}{RGB}{255,250,205}  % Light yellow
            \definecolor{color5}{RGB}{224,255,255}  % Light cyan
            \definecolor{color6}{RGB}{255,228,225}  % Light pink
            \definecolor{color7}{RGB}{245,245,220}  % Light beige
            \definecolor{color8}{RGB}{255,245,238}  % Light peach puff
            
            \node[category, fill=color1] (syntactic) at (-3, 0) {
                \textbf{Syntactic Features} \\ 
                Usage of Conjunctions \\
                Usage of Articles \\
                Frequent Usage of Function Words \\
                Usage of Personal Pronouns \\
                Usage of Pronouns \\
                Usage of Active Voice \\
                Usage of Contractions \\
                Frequent Usage of Determiners \\
                Usage of Prepositions
            };
            
            \node[category, fill=color2, below=0.25cm of syntactic] (graphical) {
                \textbf{Graphical and Digital Features} \\ 
                Usage of Numerical Substitution \\
                Usage of Uppercase Letters \\
                Usage of Text Emojis \\
                Usage of Emojis \\
                Presence of Misspelled Words \\
                Usage of Only Uppercase Letters \\
                Usage of Only Lowercase Letters \\
                Usage of Numerical Digits \\
                Frequent Usage of Punctuation
            };

            \node[category, fill=color3, right=0.5cm of syntactic] (emotional) at (-1, 0.5) {
                \textbf{Emotional and Cognitive Features} \\ 
                Positive Sentiment Expression \\
                Usage of Words Indicating Affective vs. Perceptual Processes \\
                Usage of Words Indicating Cognitive vs. Perceptual Processes \\
                Usage of Words Indicating Affective vs. Cognitive Processes \\
                Usage of Certain Tone
            };

            \node[category, fill=color5, right=0.5cm of emotional] (stylistic) {
                \textbf{Stylistic and Aesthetic Features} \\ 
                Usage of Metaphors \\
                Usage of Formal Tone \\
                Incorporation of Humor \\
                Fluency in Sentence Construction \\
                Complex Sentence Structure \\
                Usage of Sarcasm
            };
            
            \node[category, fill=color4, below=0.25cm of emotional] (social) {
                \textbf{Social and Interpersonal Features} \\ 
                Usage of Offensive Language \\
                Usage of Self-Focused Language vs. You-Focused \\
                Usage of Self-Focused Perspective vs. Third-person Singular\\
                Usage of Self-Focused Language vs. Inclusive-focused \\
                Usage of Self-Focused Language vs. Audience-focused  \\
                Usage of Polite Tone
            };

            \node[category, fill=color6, below=0.25cm of stylistic] (lexical) {
                \textbf{Lexical Features} \\ 
                Usage of Long Words \\
                Usage of Nominalizations \\
                Frequent Usage of Common Verbs
            };
            
            \node[category, fill=color7, text width=5cm] (temporal) at (5.4, -4.7) {
                \textbf{Temporal and Aspectual Features} \\ 
                Usage of Present-focused vs. Future-focused \\
                Usage of Present-focused vs. Past-focused
            };
            
            \end{tikzpicture}
        }
        \caption{We generate synthetic parallel examples to train \textsc{StyleDistance} for a wide range of style features in seven linguistic and stylistic categories. Further details on these features can be found in Appendix \ref{sec:appendix:stylefeatures}.}
        \label{fig:stylefeaturecategories}
    \end{figure*}
}

\newcommand{\umapfig}{
    \begin{figure}[H]
        \centering
        \begin{minipage}[b]{0.30\linewidth}
            \centering
            \includegraphics[width=\linewidth]{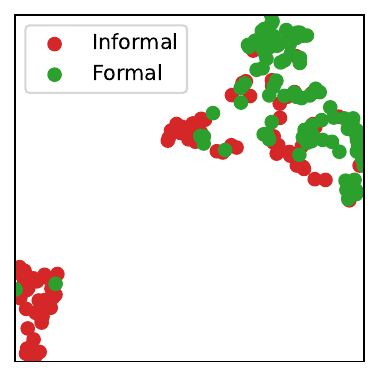}
            \caption*{\citet{styleemb}}
        \end{minipage}
        \hspace{.1\linewidth}
        \begin{minipage}[b]{0.30\linewidth}
            \centering
            \includegraphics[width=\linewidth]{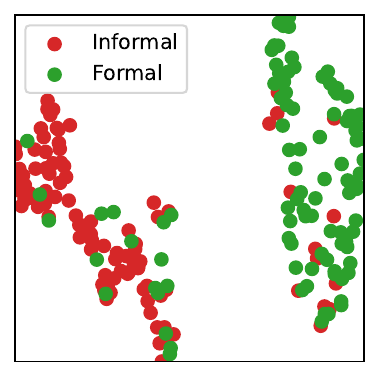}
            \caption*{\textsc{StyleDistance}}
        \end{minipage}
        \caption{UMAP visualizations of style embeddings from \citet{styleemb} and \textsc{StyleDistance} on $n=100$ random parallel formal/informal examples. \citet{styleemb} forms two distinct clusters of informal texts, making many informal examples distant in the embedding space despite sharing the same style.}
        \label{fig:umapcomparison}
    \end{figure}
}

\newcommand{\stelfig}{
    \begin{figure}[!h]
        \centering
        \includegraphics[width=215px]{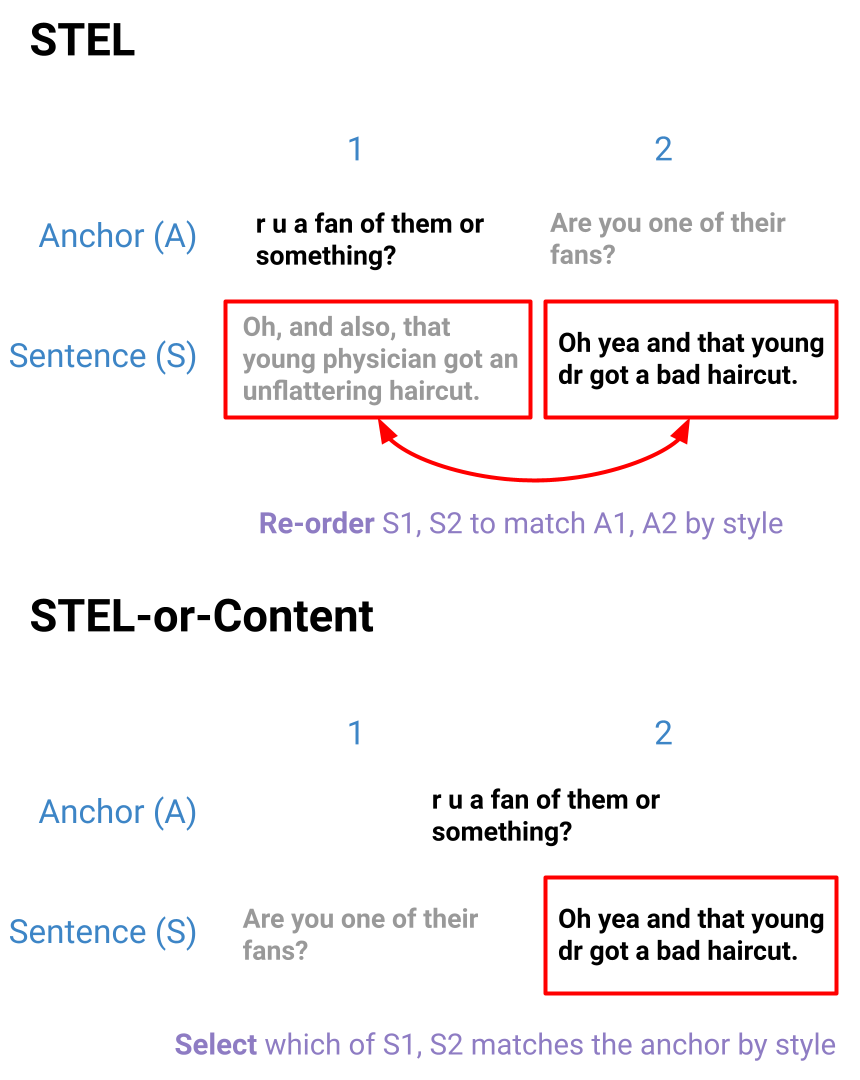}
        \caption{A visualization of the STEL and STEL-or-Content task evaluation we describe in Section \ref{sec:steleval}.}
        \label{fig:stel}
    \end{figure}
}
\newcommand{\stelevaltable}{
    \begin{table*}[!h]
    \centering
    \small
    \resizebox{\textwidth}{!}{
    \begin{tabular}{l*{12}{c}}
    \toprule[\heavyrulewidth]
    \textbf{Model} 
    & \multicolumn{2}{c}{\textbf{Formal}} 
    & \multicolumn{2}{c}{\textbf{Complex}} 
    & \multicolumn{2}{c}{\textbf{Numb3r}} 
    & \multicolumn{2}{c}{\textbf{C'tion}} 
    & \multicolumn{2}{c}{\textbf{Emoji}} 
    & \multicolumn{2}{c}{\textbf{Avg}} \\ \cmidrule(lr){2-3} \cmidrule(lr){4-5} \cmidrule(lr){6-7} \cmidrule(lr){8-9} \cmidrule(lr){10-11} \cmidrule(lr){12-13}
    & \textbf{STEL} & \textbf{S-o-C} 
    & \textbf{STEL} & \textbf{S-o-C} 
    & \textbf{STEL} & \textbf{S-o-C} 
    & \textbf{STEL} & \textbf{S-o-C} 
    & \textbf{STEL} & \textbf{S-o-C} 
    & \textbf{STEL} & \textbf{S-o-C} \\ \midrule
    \multicolumn{13}{l}{\textit{\textbf{Content-Aware Representations}}} \\ \midrule
    \texttt{roberta-base} 
    & 0.83 & 0.09 & 0.73 & 0.01 & 0.94 & 0.13 & 1.00 & 0.00 & 0.98 & 0.03 & \textbf{0.90} & \textbf{0.05} \\
    LUAR 
    & 0.80 & 0.14 & 0.67 & 0.00 & 0.74 & 0.03 & 0.77 & 0.00 & 0.99 & 0.02 & 0.86 & 0.03 \\ \midrule
    \multicolumn{13}{l}{\textbf{\textit{Content-Independent Style Representations}}} \\ \midrule
    LISA 
    & 0.73 & 0.05 & 0.65 & 0.00 & 0.85 & 0.03 & 0.92 & 0.00 & 0.77 & 0.02 & 0.71 & 0.03 \\
    \citet{styleemb} 
    & 0.83 & 0.70 & 0.58 & 0.27 & 0.56 & 0.03 & 0.96 & 0.02 & 0.95 & 0.07 & 0.77 & 0.22 \\
    \textsc{StyleDistance} 
    & 0.89 & 0.72 & 0.64 & 0.25 & 0.84 & 0.12 & 1.00 & 0.20 & 0.99 & 0.15 & \textbf{0.87} & 0.29 \\
    $\textsc{StyleDistance}_{\textsc{Synth}}$ 
    & 0.85 & 0.73 & 0.60 & 0.27 & 0.71 & 0.28 & 0.99 & 0.22 & 0.63 & 0.07 & 0.76 & \textbf{0.31} \\
    \bottomrule[\heavyrulewidth]
    \end{tabular}
    }
    \caption{Accuracy on the STEL/STEL-or-Content (S-o-C) tasks. \textsc{StyleDistance} leads on both tasks among representations trained for content-independence, and $\textsc{StyleDistance}_{\textsc{Synth}}$ generalizes remarkably well to real text data despite being trained only on synthetic data.}
    \label{table:steleval}
    \end{table*}
}
\newcommand{\synthstelevaltable}{
    \begin{table}[!t]
    \centering
    \small
    \begin{tabular}{>{\raggedright\arraybackslash}p{.3\textwidth}cc}
    \toprule
    \textbf{Model} & \textbf{STEL} & \textbf{S-o-C} \\
    \midrule
    LISA & 0.79 & 0.06 \\
    \citet{styleemb} & 0.76 & 0.25 \\
    \bottomrule
    \end{tabular}
    \caption{Results obtained by LISA and the \citet{styleemb} embeddings on STEL and STEL-or-Content instances created from the test split of our \textsc{SynthStel} dataset. 
    See Appendix \ref{sec:appendix:fullsynthstelresults} for full results.}
    \label{table:synthsteleval}
    \end{table}
}

\newcommand{\fullsynthstelevaltable}{
\begingroup
\hyphenpenalty=100000000
\exhyphenpenalty=100000000

    \begin{table*}[!h]
    \centering
    \small
    \setlength{\tabcolsep}{4pt}
    \resizebox{\textwidth}{!}{
    \begin{tabular}{>{\raggedright\arraybackslash}p{.3\textwidth}*{3}{c}c|*{4}{c}}
    \toprule
    \textbf{Style Feature} 
    & \multicolumn{2}{c}{LISA} 
    & \multicolumn{2}{c}{\citet{styleemb}} 
    & \multicolumn{2}{c}{\textsc{StyleDistance}} 
    & \multicolumn{2}{c}{$\textsc{StyleDistance}_{\textsc{Synth}}$} \\ \cmidrule(lr){2-3} \cmidrule(lr){4-5} \cmidrule(lr){6-7} \cmidrule(lr){8-9}
    & \textbf{STEL} & \textbf{S-o-C}
    & \textbf{STEL} & \multicolumn{1}{c}{\textbf{S-o-C}}
    & \textbf{STEL} & \textbf{S-o-C}
    & \textbf{STEL} & \textbf{S-o-C} \\
    \midrule
    Usage of Polite Tone & 0.93 & 0.18 & 0.76 & 0.53 & 1.00 & 1.00 & 0.78 & 1.00 \\
    Incorporation of Humor & 0.78 & 0.27 & 0.76 & 0.31 & 1.00 & 1.00 & 0.82 & 0.89 \\
    Usage of Sarcasm & 0.60 & 0.04 & 0.87 & 0.11 & 0.98 & 1.00 & 0.53 & 0.56 \\
    Usage of Metaphors & 0.91 & 0.02 & 0.53 & 0.16 & 0.98 & 0.73 & 0.58 & 0.82 \\
    Usage of Offensive Language & 1.00 & 0.40 & 0.80 & 0.13 & 1.00 & 1.00 & 0.49 & 0.93 \\
    Positive Sentiment Expression & 1.00 & 0.51 & 0.51 & 0.04 & 0.96 & 1.00 & 0.47 & 0.53 \\
    Usage of Active Voice & 0.64 & 0.00 & 0.64 & 0.07 & 1.00 & 1.00 & 0.91 & 1.00 \\
    Usage of Certain Tone & 1.00 & 0.20 & 0.60 & 0.00 & 0.87 & 1.00 & 0.87 & 0.93 \\
    Usage of Self-Focused Language vs. Inclusive-focused & 0.87 & 0.00 & 0.64 & 0.00 & 0.76 & 1.00 & 0.69 & 0.89 \\
    Usage of Self-Focused Language vs. You-Focused & 0.96 & 0.00 & 0.73 & 0.04 & 0.78 & 1.00 & 0.60 & 0.80 \\
    Usage of Self-Focused Language vs. Audience-focused & 0.73 & 0.00 & 1.00 & 0.16 & 0.91 & 1.00 & 0.67 & 1.00 \\
    Usage of Self-Focused Perspective vs. Third-person Singular & 0.76 & 0.00 & 0.44 & 0.02 & 0.73 & 1.00 & 0.60 & 0.82 \\
    Usage of Personal Pronouns & 0.64 & 0.00 & 0.56 & 0.04 & 0.82 & 1.00 & 0.80 & 0.78 \\
    Usage of Present-focused vs. Future-focused & 1.00 & 0.00 & 0.56 & 0.02 & 0.58 & 1.00 & 0.62 & 0.80 \\
    Usage of Present-focused vs. Past-focused & 0.89 & 0.00 & 0.60 & 0.04 & 0.89 & 1.00 & 0.47 & 0.91 \\
    Usage of Words Indicating Affective vs. Cognitive Processes & 1.00 & 0.07 & 0.87 & 0.09 & 0.64 & 1.00 & 0.69 & 1.00 \\
    Usage of Words Indicating Affective vs. Perceptual Processes & 0.69 & 0.02 & 0.67 & 0.00 & 1.00 & 1.00 & 1.00 & 1.00 \\
    Usage of Words Indicating Cognitive vs. Perceptual Processes & 0.76 & 0.04 & 0.56 & 0.04 & 0.91 & 1.00 & 0.71 & 0.69 \\
    Usage of Articles & 0.69 & 0.00 & 0.69 & 0.02 & 0.89 & 1.00 & 0.82 & 0.93 \\
    Fluency in Sentence Construction & 0.60 & 0.00 & 0.91 & 0.62 & 0.98 & 1.00 & 1.00 & 1.00 \\
    Frequent Usage of Function Words & 0.58 & 0.02 & 0.56 & 0.24 & 0.82 & 1.00 & 0.80 & 0.96 \\
    Frequent Usage of Common Verbs & 0.67 & 0.00 & 0.82 & 0.07 & 0.78 & 0.89 & 0.49 & 0.89 \\
    Usage of Pronouns & 0.96 & 0.00 & 0.53 & 0.24 & 0.78 & 1.00 & 0.62 & 0.80 \\
    Usage of Prepositions & 0.67 & 0.00 & 0.78 & 0.13 & 0.73 & 1.00 & 0.82 & 0.71 \\
    Frequent Usage of Determiners & 0.53 & 0.00 & 0.69 & 0.07 & 0.93 & 1.00 & 0.53 & 0.98 \\
    Usage of Conjunctions & 0.49 & 0.00 & 0.64 & 0.31 & 0.67 & 1.00 & 0.67 & 0.96 \\
    Usage of Nominalizations & 0.58 & 0.00 & 0.96 & 0.07 & 0.80 & 1.00 & 0.56 & 1.00 \\
    Usage of Long Words & 1.00 & 0.00 & 0.91 & 0.44 & 1.00 & 1.00 & 0.64 & 0.91 \\
    Usage of Numerical Digits & 0.58 & 0.00 & 0.58 & 0.02 & 0.93 & 0.80 & 0.58 & 0.62 \\
    Usage of Uppercase Letters & 0.78 & 0.00 & 1.00 & 0.98 & 1.00 & 1.00 & 1.00 & 1.00 \\
    Frequent Usage of Punctuation & 0.56 & 0.00 & 0.93 & 0.58 & 0.78 & 1.00 & 0.67 & 0.80 \\
    Usage of Formal Tone & 0.89 & 0.00 & 0.98 & 0.36 & 1.00 & 1.00 & 0.64 & 0.98 \\
    Complex Sentence Structure & 0.56 & 0.00 & 0.60 & 0.16 & 0.51 & 0.76 & 0.73 & 0.87 \\
    Usage of Contractions & 0.71 & 0.02 & 1.00 & 0.31 & 1.00 & 0.89 & 1.00 & 0.93 \\
    Usage of Numerical Substitution & 0.81 & 0.00 & 0.90 & 0.04 & 0.90 & 0.87 & 0.90 & 1.00 \\
    Usage of Only Lowercase Letters & 0.89 & 0.00 & 1.00 & 1.00 & 1.00 & 1.00 & 1.00 & 1.00 \\
    Usage of Only Uppercase Letters & 0.98 & 0.33 & 0.87 & 0.18 & 1.00 & 1.00 & 1.00 & 1.00 \\
    Presence of Misspelled Words & 1.00 & 0.24 & 1.00 & 0.91 & 1.00 & 1.00 & 1.00 & 0.96 \\
    Usage of Text Emojis & 1.00 & 0.02 & 0.91 & 0.78 & 1.00 & 1.00 & 1.00 & 1.00 \\
    Usage of Emojis & 1.00 & 0.00 & 1.00 & 0.47 & 1.00 & 1.00 & 1.00 & 1.00 \\
    \midrule
    \textbf{Average} & 0.79 & 0.06 & 0.76 & 0.25 & 0.88 & 0.97 & 0.74 & 0.89 \\
    \bottomrule
    \end{tabular}
    }
    \caption{STEL and STEL-or-Content results for top-performing models on the \textsc{SynthStel} test split. This table shows the performance variations and coverage of LISA and \citet{styleemb} embeddings for the 40 different style features found in the dataset. After training \textsc{StyleDistance} models on the \textsc{SynthStel} train split, we observe strong coverage of these 40 style features as expected, demonstrating the successful distillation of the LLM's strong style knowledge into a more efficient representation model \citep{distillation}.}
    \label{table:fullsynthsteleval}
    \end{table*}
\endgroup
}
\newcommand{\ablationtable}{
    \begin{table*}[!h]
    \centering
    \small
    \setlength{\tabcolsep}{3pt}
    \resizebox{\textwidth}{!}{
    \begin{tabular}{ll|*{12}{c}|c}
    \toprule[\heavyrulewidth]
    \multirow{2}{*}{\makecell{\textbf{Features} \\ \textbf{Tested}}} 
    & \multicolumn{1}{c}{\multirow{2}{*}{\makecell{\textbf{Features} \\ \textbf{Used}}}}
    & \multicolumn{2}{c}{\textbf{Formal}}
    & \multicolumn{2}{c}{\textbf{Complex}}
    & \multicolumn{2}{c}{\makecell{\textbf{Numb3r}}}
    & \multicolumn{2}{c}{\makecell{\textbf{C'tion}}}
    & \multicolumn{2}{c}{\textbf{Emoji}}
    & \multicolumn{2}{c}{\textbf{Avg}}
    & \multirow{2}{*}{\makecell{\textbf{Retained} \\ \textbf{Perf.}}} \\ \cmidrule(lr){3-4} \cmidrule(lr){5-6} \cmidrule(lr){7-8} \cmidrule(lr){9-10} \cmidrule(lr){11-12} \cmidrule(lr){13-14}
    & \multicolumn{1}{c}{}
    & \textbf{STEL} & \textbf{S-o-C}
    & \textbf{STEL} & \textbf{S-o-C}
    & \textbf{STEL} & \textbf{S-o-C}
    & \textbf{STEL} & \textbf{S-o-C}
    & \textbf{STEL} & \textbf{S-o-C}
    & \textbf{STEL} & \multicolumn{1}{c}{\textbf{S-o-C}}
    & \\ \midrule
    In-Domain 
    & 40 out of 40
    & 0.85 & 0.73 
    & 0.60 & 0.27 
    & 0.71 & 0.28 
    & 0.99 & 0.22 
    & 0.63 & 0.07 
    & 0.76 & 0.31 
    & 100\% \\
    Out-of-Domain 
    & 34 out of 40
    & 0.85 & 0.67 
    & 0.62 & 0.28 
    & 0.56 & 0.15 
    & 0.80 & 0.00 
    & 0.60 & 0.01 
    & 0.68 & 0.22 
    & 65\% \\
    Out-of-Distribution 
    & 25 out of 40
    & 0.64 & 0.49 
    & 0.65 & 0.26 
    & 0.57 & 0.11 
    & 0.63 & 0.02 
    & 0.68 & 0.02 
    & 0.63 & 0.18 
    & 50\% \\
    \bottomrule[\heavyrulewidth]
    \end{tabular}
    }
    \caption{We evaluate how well $\textsc{StyleDistance}_{\textsc{Synth}}$ embeddings generalize to unseen style features by ablating features from the synthetic training dataset under three  conditions: In-Domain, Out-of-Domain, Out-of-Distribution. We evaluate their performance on the STEL and STEL-or-Content (S-o-C) tasks.}
    \label{table:ablationeval}
    \end{table*}
}

\newcommand{\wegmannvsstyledistancetable}{
    \begin{figure*}[!ht]
    \setlength\dashlinedash{0.5pt} % Length of the dash
\setlength\dashlinegap{1pt}    % Gap between dashes
    \renewcommand{\arraystretch}{2} % Adds space between rows
    \centering
    \small
    \fontsize{7.5}{7.5}\selectfont
    \setlength{\tabcolsep}{8pt} % Reduces the padding between columns
    \begin{tabular}{m{.05\textwidth}m{.215\textwidth}m{.2\textwidth}m{.215\textwidth}m{.125\textwidth}}
    \toprule[\heavyrulewidth]
    & \multicolumn{2}{c}{\textbf{Contrastive example used for \citet{styleemb}}} & \multicolumn{2}{c}{\textbf{Contrastive example used for \textsc{StyleDistance} }} \\
    \noalign{\vspace{-5pt}}
    & & & \multicolumn{2}{c}{\textbf{(``Usage of Active Voice'' Style Feature) }} \\ \midrule
    \textbf{Anchor:} & \textbf{Awesome} \underline{game}. Took off from work to play it. Halfway through the week playing nonstop. Such a quality product. The \underline{Spider-Man} \underline{game} we’ve needed for years. & & \textbf{I adored} the \underline{Dolce \& Gabbana} mauve leather creation with a signature lock. & \\ \cdashline{1-5}

    \textcolor[rgb]{0.0, 0.5, 0.0}{\textbf{Positive}:} & Such a \textbf{great} \underline{Spider-Man game}. It really is the best one I’ve played. Easy to play \textbf{crazy fun} to master. There is so much you can do with \underline{Peter's} moves... & \textcolor[rgb]{0.4,0.4,0.4}{Same \textbf{Style}\newline Same \underline{Content} \newline \textit{(leads to content leakage in style representations)}} & \textbf{I observed} the impact rising temperatures had on \underline{tropical diseases}. & \textcolor[rgb]{0.4,0.4,0.4}{Same \textbf{Style}\newline Different \underline{Content}} \\

    \textcolor{red}{\textbf{Negative}:} & No \underline{Android version}? I guess \textbf{I'm not getting it} & \textcolor[rgb]{0.4,0.4,0.4}{Different \textbf{Style}\newline Different \underline{Content} \newline \textit{(leads to content leakage in style representations)}} & The \underline{Dolce \& Gabbana} mauve leather creation with a signature lock was \textbf{adored by me}. & \textcolor[rgb]{0.4,0.4,0.4}{Different \textbf{Style}\newline Same \underline{Content}} \\

    \bottomrule[\heavyrulewidth]
    \end{tabular}
    \caption{
    In the training triplets used by \citet{styleemb} (left), an anchor text is paired with a positive instance written by the same author, and a negative instance written by a different author, assuming content can be controlled via subreddit/conversation metadata. However, this assumption can fail, leading to uncontrolled content as illustrated. In our dataset used to train \textsc{StyleDistance}  (right), we control for both style and content.
    } 
    \label{figure:wegmannvsstyledistance}
    \end{figure*}
}

\newcommand{\stylefeaturestable}{
    {
    \tiny
    \renewcommand{\arraystretch}{2} % Adds space between rows
    \begin{longtable}{p{4cm} p{4cm} p{6.5cm}}
      \toprule
      \textbf{Style Feature} & \textbf{Positive and Negative Prompts} & \textbf{Style Feature Definition} \\
      \midrule
      \endfirsthead
    
      \toprule
      \textbf{Style Feature Name} & \textbf{Positive and Negative Prompts} & \textbf{Style Feature Definition} \\
      \midrule
      \endhead
    
      \bottomrule
      \endfoot
    
      \bottomrule
      \addlinespace
      \caption{The style features selected for synthetic data generation in this work.}
      \label{table:stylefeaturestable} \\
      \endlastfoot
    
Usage of Conjunctions & Positive: With conjunctions \newline Negative: Less frequent conjunctions & The "Usage of Conjunctions" text style feature refers to the use of words that connect clauses or sentences. Conjunctions are words like "and", "but", "or", "so", "because", etc. They are used to make sentences longer, more complex, or to show the relationship between different parts of a sentence. \\
Usage of Numerical Substitution & Positive: With number substitution \newline Negative: Without number substitution & Numerical substitution refers to the practice of replacing certain letters in words with numbers that visually resemble those letters. For example, replacing the letter 'e' with the number '3' in the word 'hello' to make it 'h3llo'. This is a common feature in internet slang and informal digital communication. \\
Usage of Words Indicating Affective Processes & Positive: Affective processes \newline Negative: Cognitive processes & The text style feature "Usage of Words Indicating Affective Processes" refers to the use of words that express emotions, feelings, or attitudes. These could be words that show happiness, sadness, anger, fear, surprise, or any other emotional state. The presence of such words in a text indicates that the writer is expressing some form of emotional reaction or sentiment. \\
Usage of Metaphors & Positive: With metaphor \newline Negative: Without metaphor & The "Usage of Metaphors" text style feature refers to the presence of phrases or sentences in the text that describe something by comparing it indirectly to something else. This is often done to make a description more vivid or to explain complex ideas in a more understandable way. For example, saying "time is a thief" is a metaphor because it's not literally true but it helps to convey the idea that time passes quickly and can't be regained. \\
Usage of Long Words & Positive: Long average word length \newline Negative: Short average word length & The "Usage of Long Words" text style feature refers to the frequency or prevalence of long words, typically those with more than six or seven letters, in a given text. This style feature is often used to measure the complexity or sophistication of the text. If a text has many long words, it is said to have a high usage of long words. \\
Usage of Uppercase Letters & Positive: With uppercase letters \newline Negative: Without uppercase letters & The usage of uppercase letters as a text style feature refers to the frequency or manner in which capital letters are used in a text. This could be for emphasis, to denote shouting or strong emotions, or to highlight specific words or phrases. It's not just about the start of sentences or proper nouns, but also about other uses of capital letters in the text. \\
Usage of Articles & Positive: With articles \newline Negative: Less frequent articles & The "Usage of Articles" text style feature refers to how often a text uses words like "a", "an", and "the". These words are called articles and they are used before nouns. This feature measures the frequency of these articles in a given text. \\
Usage of Text Emojis & Positive: Text Emojis \newline Negative: No Emojis & The text style feature "Usage of Text Emojis" refers to the inclusion of emoticons or smileys in the text. These are combinations of keyboard characters that represent facial expressions or emotions, such as :-D for a big grin or happy face. The presence of these symbols in a text indicates the use of this style feature. \\
Usage of Nominalizations & Positive: With nominalizations \newline Negative: Without nominalizations & Nominalizations refer to the use of verbs, adjectives, or adverbs as nouns in a sentence. This style feature is often used to make sentences more concise or formal. For example, "the investigation of the crime" is a nominalization of "investigate the crime". \\
Frequent Usage of Function Words & Positive: With function words \newline Negative: Less frequent function words & The text style feature "Frequent Usage of Function Words" refers to the regular use of words that have little meaning on their own but work in combination with other words to express grammatical relationships. These words include prepositions (like 'in', 'at', 'on'), conjunctions (like 'and', 'but', 'or'), articles (like 'a', 'an', 'the'), and pronouns (like 'he', 'they', 'it'). \\
Usage of Self-Focused Perspective or Words & Positive: Self-focused \newline Negative: Third-person singular & The "Usage of Self-Focused Perspective or Words" text style feature refers to the use of words or phrases that focus on the speaker or writer themselves. This includes the use of first-person pronouns like "I", "me", "my", "mine", and "myself", or statements that express the speaker's personal thoughts, feelings, or experiences. \\
Usage of Formal Tone & Positive: Formal \newline Negative: Informal & The "Usage of Formal Tone" text style feature refers to the use of language that is polite, impersonal and adheres to established conventions in grammar and syntax. It avoids slang, contractions, colloquialisms, and often uses more complex sentence structures. This style is typically used in professional, academic, or official communications. \\
Usage of Emojis & Positive: With Emojis \newline Negative: No Emojis & The "Usage of Emojis" text style feature refers to the inclusion of emojis, or digital icons, in a text. Emojis are often used to express emotions, ideas, or objects without using words. If a text contains emojis, it has this style feature. \\
Usage of Offensive Language & Positive: Offensive \newline Negative: Non-Offensive & The "Usage of Offensive Language" text style feature refers to the presence of words or phrases in the text that are considered rude, disrespectful, or inappropriate. These can include swear words, slurs, or any language that could be seen as insulting or derogatory. \\
Usage of Present Tense and Present-Focused Words & Positive: Present-focused \newline Negative: Future-focused & The text style feature "Usage of Present Tense and Present-Focused Words" refers to the use of verbs in the present tense and words that focus on the current moment or situation. This means the text is primarily discussing events, actions, or states that are happening now or general truths. It's like the text is talking about what is happening in the present time. \\
Presence of Misspelled Words & Positive: Sentence With a Few Misspelled Words \newline Negative: Normal Sentence & The text style feature "Presence of Misspelled Words" refers to the occurrence of words in a text that are not spelled correctly according to standard dictionary spelling. This could be due to typing errors, lack of knowledge about the correct spelling, or intentional for stylistic or informal communication purposes. \\
Incorporation of Humor & Positive: With Humor \newline Negative: Without Humor & The "Incorporation of Humor" text style feature refers to the use of language, phrases, or expressions in a text that are intended to make the reader laugh or feel amused. This could include jokes, puns, funny anecdotes, or witty remarks. It's all about adding a touch of comedy or light-heartedness to the text. \\
Usage of Personal Pronouns & Positive: With personal pronouns \newline Negative: Less frequent pronouns & The "Usage of Personal Pronouns" text style feature refers to the use of words in a text that refer to a specific person or group of people. These words include "I", "you", "he", "she", "it", "we", and "they". The presence of these words in a text can indicate a more personal or direct style of communication. \\
Fluency in Sentence Construction & Positive: Fluent sentence \newline Negative: Disfluent sentence & "Fluency in Sentence Construction" refers to the smoothness and ease with which sentences are formed and flow together. It involves using correct grammar, appropriate vocabulary, and logical connections between ideas. A text with this feature would read smoothly, without abrupt changes or awkward phrasing. \\
Usage of Only Uppercase Letters & Positive: All Upper Case \newline Negative: Proper Capitalization & The usage of only uppercase letters style feature refers to the practice of writing all the letters in a text in capital letters. This means that every single letter in the text, whether at the beginning, middle, or end of a sentence, is capitalized. It's like the 'Caps Lock' key on your keyboard is always turned on while typing the text. \\
Usage of Self-Focused Perspective or Words & Positive: Self-focused \newline Negative: Inclusive-focused & The "Usage of Self-Focused Perspective or Words" text style feature refers to the use of words or phrases that focus on the speaker or writer themselves. This includes the use of first-person pronouns like "I", "me", "my", "mine", and "myself", or statements that express the speaker's personal thoughts, feelings, or experiences. \\
Usage of Pronouns & Positive: With pronouns \newline Negative: Less frequent pronouns & The "Usage of Pronouns" text style feature refers to the frequency and types of pronouns used in a text. Pronouns are words like 'he', 'she', 'it', 'they', 'we', 'you', 'I', etc., that stand in place of names or nouns in sentences. This feature can indicate the level of personalization, formality, or perspective in a text. \\
Usage of Words Indicating Cognitive Processes & Positive: Cognitive process \newline Negative: Perceptual process & The text style feature "Usage of Words Indicating Cognitive Processes" refers to the use of words that show thinking or mental processes. These words can express understanding, knowledge, belief or doubt. For example, words like 'think', 'know', 'believe', 'understand' are used to indicate cognitive processes. \\
Complex Sentence Structure & Positive: Complex \newline Negative: Simple & The "Complex Sentence Structure" text style feature refers to sentences that contain multiple ideas or points, often connected by conjunctions (like 'and', 'but', 'or') or punctuation (like commas, semicolons). These sentences often include dependent clauses, which are parts of the sentence that can't stand alone as a complete thought, alongside independent clauses, which can stand alone. In simpler terms, if a sentence has more than one part and these parts are linked together in a way that they give more detailed information or express multiple thoughts, it has a complex sentence structure. \\
Positive Sentiment Expression & Positive: Positive \newline Negative: Negative & Positive Sentiment Expression is a text style feature that refers to the use of words, phrases, or expressions that convey a positive or optimistic viewpoint or emotion. This could include expressions of happiness, joy, excitement, love, or any other positive feelings. The text is considered to have this feature if it makes the reader feel good or positive after reading it. \\
Usage of Numerical Digits & Positive: With digits \newline Negative: Less frequent digits & The "Usage of Numerical Digits" text style feature refers to the presence and use of numbers in a text. This includes any digit from 0-9 used alone or in combination to represent quantities, dates, times, or any other numerical information. \\
Usage of Words Indicating Affective Process & Positive: Affective process \newline Negative: Perceptual process & The "Usage of Words Indicating Affective Process" text style feature refers to the use of words that express emotions, feelings, or attitudes. These words can show positive or negative sentiments, like happiness, anger, love, or hate. If a text uses a lot of these words, it means the writer is expressing a lot of emotion or personal feelings. \\
Usage of Active Voice & Positive: Active \newline Negative: Passive & The usage of active voice in a text style feature refers to sentences where the subject performs the action stated by the verb. In other words, the subject is active and directly involved in the action. For example, in the sentence "The cat chased the mouse", 'the cat' is the subject that is actively doing the chasing. \\
Usage of Only Lowercase Letters & Positive: All Lower Case \newline Negative: Proper Capitalization & The style feature "usage of only lowercase letters" refers to the practice of writing all words in a text with small letters only, without using any capital letters. This means that even the first word of a sentence, proper nouns, or the pronoun 'I' are not capitalized. It's like writing a whole text without ever pressing the shift key on your keyboard. \\
Frequent Usage of Common Verbs & Positive: With common verbs \newline Negative: Less frequent common verbs & The text style feature "Frequent Usage of Common Verbs" refers to the regular use of basic action words in a text. These are often simple, everyday verbs that are widely used in language, such as 'is', 'have', 'do', 'say', 'go', etc. If a text frequently uses these common verbs, it has this style feature. \\
Usage of Prepositions & Positive: With prepositions \newline Negative: Less frequent prepositions & The "Usage of Prepositions" text style feature refers to the use of words that link nouns, pronouns, or phrases to other words within a sentence. These words often indicate location, direction, time, or manner. Examples of prepositions include words like "in", "at", "on", "over", "under", "after", and "before". \\
Usage of Self-Focused Language & Positive: Self-focused \newline Negative: Audience-focused & The "Usage of Self-Focused Language" text style feature refers to the use of words or phrases that focus on the speaker or writer themselves. This includes the use of first-person pronouns like "I", "me", "my", "mine", and "myself". It's a way of writing or speaking where the person is often referring to their own thoughts, feelings, or experiences. \\
Usage of Certain Tone & Positive: Certain \newline Negative: Uncertain & This text style feature refers to the use of a confident tone in writing, where the author avoids using uncertain words or phrases such as 'I think', 'might', or 'seems'. This results in a text that appears more assertive and sure of the information being presented. \\
Usage of Present-Focused Tense and Words & Positive: Present-focused \newline Negative: Past-focused & The "Usage of Present-Focused Tense and Words" text style feature refers to the use of verbs in the present tense and words that focus on the current moment or situation. This means the text is primarily discussing events, actions, or states that are happening right now or generally true. \\
Usage of Sarcasm & Positive: With sarcasm \newline Negative: Without sarcasm & The "Usage of Sarcasm" text style feature refers to the presence of statements or expressions in the text that mean the opposite of what they literally say, often used to mock or show irritation. This style is often characterized by irony, ridicule, or mockery, and is used to express contempt or to criticize something or someone in a humorous way. \\
Usage of Self-Focused Perspective or Words & Positive: Self-focused \newline Negative: You-focused & The "Usage of Self-Focused Perspective or Words" text style feature refers to the use of words or phrases that focus on the speaker or writer themselves. This includes the use of first-person pronouns like "I", "me", "my", "mine", and "myself", or statements that express the speaker's personal thoughts, feelings, or experiences. \\
Frequent Usage of Punctuation & Positive: With frequent punctuation \newline Negative: Less Frequent punctuation & The text style feature "Frequent Usage of Punctuation" refers to the regular and abundant use of punctuation marks such as commas, periods, exclamation points, question marks, etc., in a piece of text. This style feature is present when the writer often uses these symbols to structure their sentences, express emotions, or emphasize certain points. \\
Usage of Polite Tone & Positive: Polite \newline Negative: Impolite & The "Usage of Polite Tone" text style feature refers to the use of respectful and considerate language in a text. This can include using words like 'please', 'thank you', or phrases that show deference or respect to the reader. It's about making the text sound courteous and respectful, rather than demanding or rude. \\
Usage of Contractions & Positive: With contractions \newline Negative: Without contractions & The "Usage of Contractions" text style feature refers to the use of shortened forms of words or phrases in a text. These are typically formed by omitting certain letters or sounds and replacing them with an apostrophe, such as "don't" for "do not" or "I'm" for "I am". If a text frequently uses such shortened forms, it has this style feature. \\
Frequent Usage of Determiners & Positive: With determiners \newline Negative: Less frequent determiners & The text style feature "Frequent Usage of Determiners" refers to the regular use of words that introduce a noun and give information about its quantity, proximity, definiteness, etc. These words include 'the', 'a', 'an', 'this', 'that', 'these', 'those', 'my', 'your', 'his', 'her', 'its', 'our', 'their'. If a text often uses such words, it has this style feature. \\

    \end{longtable}
    }
}
\newcommand{\trainingdetailstable}{
    \begin{table}[H]
    \small
    \centering
    \begin{tabular}{m{.62\textwidth}>{\centering\arraybackslash}m{.3\textwidth}}
    \toprule[\heavyrulewidth]
    \textbf{Hyperparameter} & \textbf{Value} \\ \midrule
    Model & \texttt{Facebook/roberta-base} \\
    Hardware & 4x or 8x NVIDIA RTX A6000 \\
    Distributed Protocol & PyTorch FSDP \\
    Data Type & \small{\texttt{torch.bfloat16}} \\
    Loss Function & \texttt{TripletLoss} \citep{tripletloss} \\
    Triplet Loss Margin & 0.1 \\
    LoRA \citep{lora} & \small{\texttt{all-linear, r=8\newline \hspace{4em} lora\_alpha=8 \newline lora\_dropout=0.0}} \\
    Optimizer & \texttt{adamw\_torch} \\
    Learning Rate & 1e-4 \\
    Weight Decay & 0.01 \\
    Learning Rate Scheduler & \small{\texttt{linear}} \\
    Warmup Steps & 0 \\
    Batch Size & 512 \\
    Train-Validation Split & 90/10\% \\
    Early Stopping Threshold & 0.0 \\
    Early Stopping Patience & 1 epoch \\

    \bottomrule[\heavyrulewidth]
    \end{tabular}
    \caption{Hyperparameters selected for contrastive learning training experiments.}
    \label{table:hyperparamtable}
    \end{table}
}
\newcommand{\attributedprompttable}{
    \begin{table}[H]
    \small
    \fontsize{8}{9}\selectfont
    \centering
    \begin{tabular}{m{.30\columnwidth} >{\raggedright\arraybackslash}p{.50\columnwidth}}
    \toprule[\heavyrulewidth]
    \textbf{Attribute} & \textbf{Values} \\ \midrule
    \textbf{Topic} & \textit{A randomly extracted fine-grained topic from C4.} \\
    \textbf{Sentence Length} & \texttt{[`10-15 words', `15-20 words', `20-25 words', `25-30 words']} \\
    \textbf{Point of View} & \texttt{[`first-person', `second-person', `third-person']} \\
    \textbf{Tense} & \texttt{[`past', `present', `future']} \\
    \textbf{Type of Sentence} & \texttt{[`Declarative', `Semicolon Structure (compound)', `Question', `Exclamation']} \\
    \bottomrule[\heavyrulewidth]
    \end{tabular}
    \caption{Attributes sampled for the attributed prompt string generation.}
    \label{table:attributedprompttable}
    \end{table}
}

\newcommand{\avevaltable}{
    \begin{table}[t!]
    \centering
    \fontsize{10}{11}\selectfont
    \setlength{\tabcolsep}{2.5pt}
    \scalebox{0.85}{
    \begin{tabular}{>{\raggedright\arraybackslash}p{2.5cm}cccccc}
    \toprule[\heavyrulewidth]
    \textbf{Model} & \textbf{PAN'11} & \textbf{PAN'13} & \textbf{PAN'14} & \textbf{PAN'15} & \textbf{Avg} \\ \midrule
    LISA & 0.55 & 0.66 & 0.43 & 0.64 & 0.57 \\
    \citet{styleemb} & 0.65 & 0.39 & 0.57 & 0.63 & 0.56 \\
    \textsc{StyleDistance} & 0.62 & 0.68 & 0.58 & 0.67 & \textbf{0.64} \\
    \bottomrule[\heavyrulewidth]
    \end{tabular}}
    \caption{ROC-AUC results on the PAN 2011-2015 Authorship Verification (AV) shared tasks.}
    \label{table:avevaltable}
    \end{table}
}
\newcommand{\styletransferevaltable}{
    \begin{table}[!t]
    \centering
    \small
    \fontsize{10}{11}\selectfont
    \begin{tabular}{l c}
    \toprule[\heavyrulewidth]
    \textbf{Model} & \textbf{Accuracy} \\ \midrule
    Human & 0.37 \\
    \midrule
    LUAR & 0.38 \\
    \citet{styleemb} & 0.39 \\
    \textsc{StyleDistance} & \textbf{0.46} \\
    \bottomrule[\heavyrulewidth]
    \end{tabular}
    \caption{Accuracy results on style transfer evaluation.}
    \label{table:styletransfereval}
    \end{table}
}
\newcommand{\datasetevaltable}{
    \begin{table}[!h]
    \renewcommand{\arraystretch}{1.4} % Adds space between rows
    \centering
    \small
    \fontsize{10}{11}\selectfont
    \setlength{\tabcolsep}{5pt}
    \begin{tabular}{lcc}
    \toprule[\heavyrulewidth]
    \textbf{Metric} & \textbf{Baseline} & \textbf{Score} \\ \midrule
    \makecell{\textbf{Style Feature Presence}~~~~~~ \\ \textit{(\% humans judged correct)}} & 0.50 & 0.92 \\
    \textbf{Content Similarity} & 0.88 & 0.88 \\
    \textbf{Fluency} & 0.80 & 0.92 \\
    \textbf{Diversity} & 0.95 & 0.91 \\
    \bottomrule[\heavyrulewidth]
    \end{tabular}
    \caption{Results of the human and automatic evaluations of our synthetic dataset.}%Human and automatic evaluation metrics of the synthetic dataset.}
    \label{table:dataseteval}
    \end{table}
}
\newcommand{\smalltinystyleroutputs}{
    \begin{table}[!h]
    \setlength\dashlinedash{0.5pt} % Length of the dash
    \setlength\dashlinegap{1pt}    % Gap between dashes
    \small
    \fontsize{8}{9}\selectfont
    \centering
    \begin{tabular}{m{.35\columnwidth} >{\raggedright\arraybackslash}p{.45\columnwidth}}
    \toprule[\heavyrulewidth]
    \textbf{Source Text (Informal)} & \textit{"its keeping me up at nite, i have to know what it is"} \\ \cdashline{1-2} \rule{0pt}{10pt}
    \citet{styleemb} \textbf{\newline \textcolor{white}{--} ($\rightarrow$ Formal)} & \textit{"Have to know what this is, keeping me up at night."} \\ 
    \textbf{\textsc{StyleDistance} \newline \textcolor{white}{--} 
 ($\rightarrow$ Formal)} & \textit{"What is it? It is keeping me up at night."} \\
    \bottomrule[\heavyrulewidth]
    \end{tabular}
    \caption{A demonstration of TinyStyler conditioned on \textsc{StyleDistance} embeddings.}
    \label{table:smalltinystyleroutputs}
    \end{table}
}

\newcommand{\tinystyleroutputs}{
    {
    \small

    \renewcommand{\arraystretch}{2} % Adds space between rows
\begin{longtable}{p{4.5cm} p{4.5cm} p{4.5cm}}
    \toprule
    \textbf{Source Text (Informal)} & \textbf{\textsc{Wegmann} ($\rightarrow$ Formal)} & \textbf{\textsc{StyleDistance} ($\rightarrow$ Formal)} \\
    \midrule
    \endfirsthead
    \bottomrule
    \endfoot
    
    \bottomrule
    \addlinespace
    \caption{Outputs from TinyStyler, conditioned on Wegmann embeddings and \textsc{StyleDistance} embeddings.}
    \label{table:tinystyler-outputs} \\
    \endlastfoot
    \textit{"its keeping me up at nite, i have to know what it is"} & \textit{"Have to know what this is, keeping me up at night."} & \textit{"What is it? It is keeping me up at night."} \\
    \textit{"i like journey's open arms... but i like mariah carey's version of that song better."} & \textit{"I prefer mariah carey's version of Journey, Open Arms."} & \textit{"I think mariah carey's version of that song is better."} \\
    \textit{"And you can't rely on rumors."} & \textit{"You can never trust rumors about it."} & \textit{"I would say that you can't rely on rumors."} \\
    \midrule
 \textbf{Source Text (Formal)} & \textbf{\textsc{Wegmann} ($\rightarrow$ Informal)} & \textbf{\textsc{StyleDistance} ($\rightarrow$ Informal)} \\
    \midrule

    \textit{"I favor the man as he is humorous and grounded."} & \textit{"i love tahm as he is humorous and grounded..."} & \textit{"i'm gonna go with the man...hes humorous and grounded."} \\
    \textit{"I am sure you will both enjoy it."} & \textit{"ok i am sure you both will enjoy it!"} & \textit{"oh yea...you'll both enjoy it"} \\
    \textit{"I very much enjoy this song."} & \textit{"i love this song so much, i'm a fan of it."} & \textit{"i like this song so much man..."} \\

\end{longtable}

    }
}
\usepackage[preprint]{acl}
\usepackage{times}
\usepackage{latexsym}
\usepackage{arydshln}
\usepackage[T1]{fontenc}
\usepackage[utf8]{inputenc}

\usepackage{microtype}

\usepackage{inconsolata}

\usepackage{graphicx}

\title{\textsc{StyleDistance}: Stronger Content-Independent Style Embeddings \\ with Synthetic Parallel Examples}

\author{%
  Ajay Patel \textsuperscript{\textdagger~}\thanks{Denotes equal contribution; direct correspondence to: \href{mailto:ajayp@upenn.edu}{ajayp@upenn.edu}}~~~~~~~~~Jiacheng Zhu\textsuperscript{\textdagger~}\footnotemark[1]~~~~~~~~~Justin Qiu\textsuperscript{{\textdagger}}~~~~~~~~~Zachary Horvitz\textsuperscript{{\textdaggerdbl}}   \\\textbf{Marianna Apidianaki\textsuperscript{\textdagger}~~~~~~~~~Kathleen McKeown\textsuperscript{\textdaggerdbl}~~~~~~~~~Chris Callison-Burch\textsuperscript{\textdagger}} \\
  University of Pennsylvania\textsuperscript{\textdagger}~~~~~~~~~Columbia University\textsuperscript{\textdaggerdbl}
}
\begin{document}
\maketitle

% Main sections
\begin{abstract}
Style representations aim to embed texts with similar writing styles closely and texts with different styles far apart, regardless of content. However, the contrastive triplets often used for training these representations may vary in both style and content, leading to potential content leakage in the representations. We introduce \textsc{StyleDistance}, a novel approach to training stronger content-independent style embeddings. We use a large language model to create a synthetic dataset of near-exact paraphrases with controlled style variations, and produce positive and negative examples across 40 distinct style features for precise contrastive learning. We assess the quality of our synthetic data and embeddings through human and automatic evaluations.  \textsc{StyleDistance} enhances the content-independence of style embeddings, which generalize to real-world benchmarks and outperform leading style representations in downstream applications. Our model can be found at \url{https://huggingface.co/StyleDistance/styledistance}.
\end{abstract}
\mainfig

\section{Introduction}
\label{sec:introduction}

The most common objective when training text embeddings is to place texts with similar semantics close together in the embedding space \citep{sentencetransformers}. Style representations, by contrast, aim to embed texts with similar writing styles near each other and texts with different styles far apart, regardless of their semantic content \citep{styleemb}. Embeddings are usually trained via contrastive learning, with triplets consisting of an anchor text, a positive text (which should be embedded closely to the anchor), and a negative text (which should be embedded far from the anchor) \citep{contrastive1, contrastive2, tripletloss}. Existing approaches often use social media datasets with the assumption that all writing by the same author shares a similar style and that texts by different authors exhibit dissimilar styles \citep{styleemb}. These methods also attempt to minimize content representation in the resulting embeddings. They select a text from the same author on a \textit{different topic} as the positive example, and a text from a different author on the \textit{same topic} as the negative example, approximating topic similarity using subreddit or conversation metadata. However, these methods are limited by the imperfect nature of data acquired under such assumptions. For example, the same author may write about the same topic even in different subreddits. As a result, these imperfect contrastive triplets do not explicitly control for content, leading to style embeddings with weak content-independence.
%\textcolor{gray}{A solution to this problem would be to use parallel datasets where style varies while content is held constant, but such datasets are rare. Without parallel data, training must instead rely on proxy objectives, such as using authorship-based contrastive triplets, which can introduce content leakage into style representations (see Figure \ref{figure:wegmannvsstyledistance}).} 
The ``content leakage'' caused by such proxy objectives (illustrated in Figure \ref{figure:wegmannvsstyledistance}) can undermine the effectiveness of style representations in tasks that require strict separation between style and content, such as stylistic analysis, authorship tasks, style transfer steering, and automatic style transfer evaluation. To overcome this, a more controlled approach to style contrastive learning is necessary.

In this paper, we introduce \textsc{StyleDistance}, a novel method for training stronger, content-independent style embeddings which leverages synthetic parallel text examples generated by a large language model (LLM) \citep{gpt4}. By creating near-exact paraphrases with controlled stylistic variations, we produce positive and negative examples across 40 distinct style features. This synthetic dataset, which we call \textsc{SynthStel}, enables more precise contrastive learning (visualized in Figure \ref{fig:main}) and is more robust to the content leakage inherent in existing datasets. We evaluate our method on both human and automated benchmarks, measuring the content-independence, quality, and utility of \textsc{StyleDistance} embeddings.

In summary, our primary contributions are:

\begin{enumerate}
\item We generate and release \textsc{SynthStel}, a dataset of near-exact paraphrases across 40 distinct style features.

\item We introduce \textsc{StyleDistance}, a new approach to style representation learning which uses synthetic parallel examples with controlled stylistic variations. We release the embedding model as a resource.

\item We demonstrate that \textsc{StyleDistance} significantly improves the content-independence of style embeddings, generalizing effectively to real-world benchmarks of style representation quality and outperforming existing style representations in downstream applications.
\end{enumerate}
\wegmannvsstyledistancetable
\section{Related Work}
\label{sec:relatedwork}

\paragraph{Style Representations} In previous work, style representations were learned from unlabeled texts \citep[inter alia]{styletransfervector2,styletransfervector6,textsettr}. Due to the lack of parallel datasets covering diverse style features, \citet{deepstyle} and \citet{styleemb} have both recently used authorship as a proxy for style as we discussed previously. \citet{luar} trained
embeddings to uniquely represent different authors. Although these embeddings capture  features representative of authors' style, they also capture content features due to the lack of control for content-related aspects. \citet{lisa}  trained LISA, a style vector created by annotating texts with their stylistic features using LLMs. LISA, however, trades off performance to create a vector with interpretable dimensions. The highest quality style representations to date come from approaches that employ a contrastive learning objective \citep{styleemb}. In this paper, we propose leveraging the strengths of generative LLMs to build a dataset that will serve to train strong content-independent style embeddings using a contrastive learning objective.  From a practical perspective, style representations are useful in downstream applications such as arbitrary text style transfer \citep{learningtogenerate,paraguide,tinystyler} where they help steer and guide transfer, and content-independence is important for reducing hallucinations in generations.

\paragraph{LLMs and Text Style} LLMs are commonly used for style transfer \citep{reif,promptandrerank,lowresourceauthorshipstyletransfer,styleremix} and style analysis \citep{iclef,lisa}. Their strength with style-related tasks has been demonstrated, yet leveraging this knowledge to build strong content-independent style representations has not yet been explored.

\section{Data Generation}
\label{sec:datageneration}

A core component of our proposed {\sc StyleDistance} approach is a LLM that generates a synthetic dataset with controlled content and style. The dataset is composed of pairs of sentences which are paraphrases of each other (i.e. their content is similar) but differ in style. Each pair includes a positive example that showcases a specific stylistic feature (e.g., usage of active voice or emojis) and a corresponding negative example that lacks this feature. In this section, we outline the selected style features and provide implementation details for our synthetic data generation procedure. Additionally, we present evaluations conducted to assess the quality of the synthetic dataset prior to its use in training our style embeddings.

\paragraph{Style Feature Selection} There is no predefined set of style features, and what features are considered to describe style vs. content can vary across different studies \citep{tstsurvey}. For this work, we select 40 style features across 7 broad categories (visualized in Figure \ref{fig:stylefeaturecategories}) which have been addressed in different  works on text style \citep{liwc,hovy,stel,tstsurvey,lisa}. Specifically, we select features for which it is possible to generate both positive and negative examples (e.g., formal/informal, passive/active voice). Since some  features can blur the line between style and content (e.g., usage of sarcasm), it might be difficult to generate perfectly parallel positive and negative pairs, with the same content. For these features, we control the  generation as much as possible with the aim to obtain near-exact paraphrases. Furthermore, some style features may be impossible to fully remove from a sentence in order to generate a negative example (e.g., usage of articles). For these, we aim for the positive example to contain the feature with higher frequency than the negative example. For more details on all the selected style features, see Appendix \ref{sec:appendix:stylefeatures}. While these 40 features may not cover the infinite number of styles that may exist, we believe they can serve to learn more primitive features (e.g., use of contractions, use of long words, use of formal style) which may help generalize to more complex styles  involving these features (e.g., ``professorial style''). We discuss and test this generalization assumption in Section \ref{sec:steleval}.

\stylefeaturecategoriesfig

\paragraph{Generation} % For the synthetic dataset,
For each of the selected  features, we generate 100 pairs of positive and negative examples  by prompting GPT-4 \citep{gpt4} with the DataDreamer library \citep{datadreamer}. %to generate 
Each generated pair contains a sentence where the feature is present (positive example), and a paraphrase where the feature is absent or less present (negative example). 
%``a pair of \{\texttt{positive\_feature}\} and \{\texttt{negative\_feature}\} sentences''.

\attributedprompttable

\citet{attrprompt} found that LLMs struggle with diversity when prompted to generate text examples.  We use their proposed attributed prompt (AttrPrompt) method to ensure generations are sufficiently diverse and varied across basic attributes, such as ``Sentence Length'' and ``Type of Sentence''. The method randomly selects values from a defined set of attributes to be included in the prompt, which serve as conditioning for the text generation. In Table \ref{table:attributedprompttable}, we showcase the attributes we sample from in our attributed prompt in order to vary our generations. See Appendix \ref{sec:genposandneg} for details on our full attributed prompt and inference parameters.

For the ``Topic'' attribute, we sample fine-grained distinct topics for each generation %, we extract fine-grained topics 
from the C4 corpus \citep{t5andc4}. We do this by extracting a random sentence from a random document in C4, and we then use a zero-shot prompt (given in Appendix \ref{sec:extracttopic}) with GPT-4 to identify the fine-grained topic of that sentence. We employ several heuristics to select sentences from C4 that have desired characteristics: written in English, sufficiently long (greater than 32 words), and consist of natural text rather than formatting text found in some C4 documents. We provide an implementation of these heuristics in our supplementary materials.

We call our final synthetically generated dataset \textsc{SynthStel},  and create train and test splits using a 90\%/10\% split stratified by style feature.

\paragraph{Dataset Evaluation}

%We evaluate 
We conduct a human and an automatic evaluation of the quality of our synthetic dataset for a number of different properties using the test split. % with human and automatic evaluation on the test split.

First, we measured the extent to which humans judge our positive examples do contain the desired style feature  and our negative examples do not. Note that the appreciation of some style features  (such as ``Incorporation of Humor'') can be subjective, and some other features (e.g., ``Usage of Articles'') %can only be increased or decreased in frequency but 
are not fully removed in negative examples but might appear less frequently therein than in positive examples. In spite of these intricacies which made the annotators' task more difficult, our human evaluation results were strong: 92\% of the time, annotators judged the positive and negative labels correct (random chance is 50\%).
%with the generated positive and negative examples. 
Each % task 
instance was annotated by 10 different annotators from a pool of 73  % distinct human annoators drawn from a population of  
graduate students in a NLP class. %taking a class on natural language processing. 
We also assessed inter-annotator agreement with Krippendorf's Alpha \citep{krippendorffalpha} and achieved a reliability score of 0.55. For more details about the human annotation, see Appendix \ref{sec:appendix:humanannotation}. 

We also run automatic evaluations to assess other properties of the dataset: 
\begin{itemize}
    \setlength\itemsep{1pt}
    \item \textbf{Content Similarity} measures the average semantic similarity between the positive and negative parallel examples \citep{sentencetransformers}.\footnote{For % any 
    evaluations using semantic similarity, we use the \texttt{all-mpnet-base-v2} model.}
\item \textbf{Fluency} measures the average fluency of our examples\footnote{We exclude generations for style features that specifically address disfluency.} %are intentionally meant to be disfluent.} 
using a classifier trained on CoLA \citep{cola}. 
\item \textbf{Diversity} uses the % diversity measurement score in 
score proposed by \citet{diversityscore} to measure %which measures 
how different each generated text % we generate 
is from every other in terms of content/topic using semantic similarity.%, as shown. in equation \eqref{eq:diversity}% text using semantic similarity to measure diversity in content / topic:

% \vspace{-5mm}
% \begin{equation}
%     \label{eq:diversity} 
% \scalebox{0.75}{
% $
% \operatorname{div}(T)=\frac{1}{|T|^2-|T|} \sum_{t_i \in T} \sum_{t_j \in T}^{i \neq j}\left(1-\operatorname{sim}\left(t_i, t_j\right)\right)
% $
% }
% \end{equation}

\end{itemize}

 We compute a baseline for these scores with natural data from the dataset of sentence pairs in \citet{stel}. The results of these evaluations can be found in Table \ref{table:dataseteval}. Our generated examples fare well in all these aspects: they are topically diverse and fluent, and the similarity inside each pair of positive/negative examples is high. An additional (less direct) evaluation of the quality of our synthetic dataset is proposed in Section \ref{sec:evaluation}, where we evaluate the \textsc{StyleDistance} embeddings that we train on this dataset.

\datasetevaltable
\section{\textsc{StyleDistance}}
\label{sec:styledistance}

We next describe how we trained \textsc{StyleDistance} embeddings using our synthetic dataset.

\subsection{Sampling Contrastive Triplets}
\label{sec:triplets}

After generating 100 pairs of positive and negative examples for each style feature, we construct feature-specific triplets as follows:  %triplets. % by iterating over each feature. % For each example, 
 We select an ``anchor'' ($a$) and a ``positive'' example ($p$) %({\tt pos}) 
 from different pairs available for a % within the same 
feature, ensuring that the two examples are {\bf identical in style}  but not in content. For the ``Usage of Active Voice'' example in Figure \ref{fig:main}, % the anchor and {\tt pos} 
$a$ and $p$ are two active sentences (\textit{I adored ...}, \textit{I observed...}) on  different topics. 
As a negative example ($n$), we use 
% is randomly sampled either from the same row as the anchor or from the positive row. 
the paraphrase of either $a$ or $p$   % the anchor or the {\tt pos} %positive 
%example 
which does not contain the feature; therefore, $n$ %negative example 
is  always different in style. % from %a different style than 
%both the anchor and {\tt pos}. 
In the example in Figure \ref{fig:main}, $n$ is the paraphrase of the anchor in passive voice (\textit{...adored by me}). 

In terms of {\bf content}, $n$ % {\tt neg} %the negative example is parallel to 
is a paraphrase of the anchor $a$ in half of the triplets, and  has different content in the other half. % it is different from the anchor %it differs from it in both style and content.} 
%and is a parallel example to the anchor in half the triplets and, in the other half, the negative example is different in both style and content. 
This ensures that the trained model will not only learn to % the embedding does not overfit to only being able to discriminate perfectly 
discriminate parallel (paraphrased) texts, but will be able to generalize to texts with different content during inference. This sampling process results in \textasciitilde 320K unique triplets. The implementation of this simple algorithm is shared in the supplementary materials.

\subsection{Contrastive Learning Objective}

In our final dataset of triplets $\mathcal{D}$, each triplet $(a, p, n) \in \mathcal{D}$ contains an anchor text ($a$), a positive text ($p$), and a negative text ($n$). We train our embedding model $f_{\theta}(\cdot)$ with a triplet loss (with margin $\alpha$) \citep{tripletloss}:
\[
\scalebox{0.72}{
    $L_{\text{t}}(\theta) = \sum_{(a, p, n) \in \mathcal{D}} \left[\left\|f_\theta(a) - f_\theta(p)\right\|_2^2 - \left\|f_\theta(a) - f_\theta(n)\right\|_2^2 + \alpha\right]_{+}$
}
\]

\noindent We use \texttt{roberta-base} as our base model for fine-tuning---the same base model used for the style embeddings in \citet{styleemb}---and perform training with DataDreamer and LoRA \citep{datadreamer,lora}. We use a margin of 0.1, a learning rate of 1e-4, and a batch size of 512. We further split our training set % split 
into a train and validation split (90\%/10\%) and train using an early stopping patience of 1 epoch. For full details on the training setup, see Appendix \ref{sec:appendix:trainingdetails}.

\stelevaltable
\ablationtable

\subsection{Training}

We train two versions of our model. $\textsc{StyleDistance}_{\textsc{Synth}}$ is fine-tuned only on the synthetic triplets described in Section \ref{sec:triplets}. We also train a version % where we test the effect of 
using the synthetic triplets for %as
data augmentation ($\textsc{StyleDistance}$). In this case, our training set is comprised of % where we use 
50\% natural data---i.e. the triplets used to train the \citet{styleemb} model\footnote{We use the \texttt{train-conversation} split.}--- and 50\% synthetic data. For the augmented model, we hypothesize that mixing in these perfectly parallel synthetic examples will help regularize the model, and discourage it from representing content-related features in favor of style-related ones offering the potential advantages of both approaches: (1) enhanced content-independence, and (2) the ability to capture niche style features in the natural data. We provide a visualization of the learned embedding space of our model after contrastive training using UMAP \citep{umap} in Appendix \ref{sec:appendix:umap}. %\citep{umap}

\section{Evaluation}
\label{sec:evaluation}

We propose a direct evaluation of the quality of \textsc{StyleDistance} embeddings, and an evaluation of their utility in downstream applications. We use other leading style representations like LISA \citep{lisa} and the embeddings from \citet{styleemb} as baselines, and compare them with our style embeddings on the STEL and STEL-or-Content tasks \citep{stel,styleemb}. We also compare to the LUAR model \citep{luar}, an authorship representation model which does not explicitly train content-independent representations but captures both content and style features in an attempt to represent different authors. LUAR is sometimes used for automatic style transfer evaluation, where we believe a strong style representation would be better suited. We thus choose this task for evaluation. % Finally, we evaluate whether our embeddings can be used to guide arbitrary text style transfer with the \textsc{TinyStyler} \citep{tinystyler} model.

\subsection{STEL and STEL-or-Content Evaluation}
\label{sec:steleval}

We first benchmark our embeddings on the STEL and STEL-or-Content tasks that allow for a direct evaluation of the quality of style representations. We briefly describe these tasks below and illustrate examples of these tasks in Appendix \ref{sec:appendix:stelfig}:

\begin{itemize}

\item \textbf{STEL:} Given two anchor sentences (A1, A2) and two test sentences (S1, S2), STEL measures the ability of an embedding model to pair each test sentence with the anchor sentence that shares the same style based on the cosine similarity of the embeddings of the sentences.

\item \textbf{STEL-or-Content:} The STEL-or-Content task is similar to STEL but more adversarially challenging, hence  better for testing content-independence. In this task, there are again two test sentences (S1, S2) but only a single anchor sentence (A). The test sentence that best matches the \emph{style} of the anchor must be selected; but the incorrect test sentence---which is written in a different style---is a  paraphrase of the anchor with similar \emph{content}. Therefore, in order to succeed on the STEL-or-Content task, a model needs to represent style features stronger than content features.

\end{itemize}

In their paper, \citet{stel} provide a STEL and STEL-or-Content evaluation benchmark over five features with curated natural data. We test our models on this benchmark and present the results in Table \ref{table:steleval}. Our results are consistent with results reported by \citet{stel}, who showed that even untrained models like \texttt{roberta-base} can capture style information well, resulting in stronger STEL performance than any of their fine-tuned models. However, the more challenging STEL-or-Content task, which better tests content-independence, shows that only models specifically trained for content-independence are able to capture style features better than content features. Our results indicate that the embeddings generated with our \textsc{StyleDistance} approach lead  over other style representations on both the STEL and STEL-or-Content tasks. Interestingly, we find that $\textsc{StyleDistance}_{\textsc{Synth}}$ captures style remarkably well, and manages to generalize to the natural text examples in the evaluation benchmark despite only being trained on our synthetic contrastive triplets. We conclude using synthetic parallel examples during training makes the model more content-independent and helps better capture style. 

\subsection{Generalization Experiment} 

In an ablation study, we test the ability of our training approach to produce embeddings that can generalize to unseen style features which are not present in the synthetic dataset. We conduct this evaluation by ablating features from the data used to train $\textsc{StyleDistance}_{\textsc{Synth}}$ under three conditions and show the results in Table \ref{table:ablationeval}. In the \textbf{In-Domain} condition, all 40 style features are included. In the \textbf{Out-of-Domain} condition, we exclude synthetic examples corresponding to the five style features in the STEL/STEL-or-Content benchmark \footnote{In our 40 features, there are two separate features for emoji and text emoticons (:-D) so we exclude 6 total features for this condition instead of 5 features, resulting in 34 features.}. In the \textbf{Out-of-Distribution} condition, we further exclude examples for any features similar or indirectly related to the five evaluated features. Details on the exact 15 style features ablated can be found in Appendix \ref{sec:appendix:ablationdetails}. We compare the Out-of-Domain and Out-of-Distribution performance of $\textsc{StyleDistance}_{\textsc{Synth}}$ on the two tasks to its In-Domain performance, obtained when it was trained on data for all 40 style features. Even in the challenging Out-of-Distribution condition, $\textsc{StyleDistance}_{\textsc{Synth}}$ retains 50\% of its performance on the challenging STEL-or-Content task (see the ``Retained Perf.'' column). This study indicates our training approach generalizes reasonably well to out-of-domain style features which can be composed from style features selected for generation and, to some extent, even to out-of-distribution style features fully outside the selected set.

\subsection{Synthetic Data for Probing}

Our previous experiments demonstrate that training on our synthetic dataset yields strong style embeddings. Next, we investigate whether the synthetic dataset can be used for an entirely different purpose: to probe which specific style features are captured by existing style representations. Our  \textsc{SynthStel} dataset allows for the creation of synthetic STEL and STEL-or-Content task instances across a  range of 40 style features---much broader than the \citet{stel} benchmark where five features were addressed. We use the test split of \textsc{SynthStel} to generate task instances for probing. We examine whether LISA vectors and the \citet{styleemb} style embeddings capture these 40 style features, which \textsc{StyleDistance} models are directly trained to represent. We show average results over all 40 features in Table \ref{table:synthsteleval}. Per feature results are provided in Appendix \ref{sec:appendix:fullsynthstelresults}. Our findings reveal only moderate coverage by LISA and \citet{styleemb}, with high variance depending on the evaluated feature (e.g., ``Usage of Nominalizations'' is poorly captured, with a near-zero STEL-or-Content score for both models). We calculate the mean squared error (MSE) between the STEL and STEL-or-Content scores for the real and synthetic task instances across the five features in the real benchmark, finding an average MSE of 0.039. This small MSE value shows that using synthetic data for probing can reasonably serve to assess which style features are represented by a model without need for manual example curation.

\synthstelevaltable

\subsection{Downstream Evaluation}

We next evaluate and/or demonstrate our style embeddings in three downstream applications.

\paragraph{Authorship Verification} We first test our style embeddings on the authorship verification (AV) task \citep{authorshipverification}. 
Given two documents by unknown authors, the goal of the authorship verification (AV) task is to determine whether they were written by the same author, based on their stylistic similarities and differences \citep{stylisticauthorshipverification}. We use a series of AV shared task datasets released by PAN in 2011-2015 \citep{pan11,pan13,pan14,pan15}\footnote{PAN is the ``Plagiarism Analysis, Authorship Identification, and Near-Duplicate Detection'' workshop. No AV shared task was proposed in 2012. (URL: \url{https://pan.webis.de})}. Since the two documents may be about different topics, good content-independent style representations would be expected to perform better in the AV task than embeddings that capture content. We calculate the cosine similarity of the two documents using our tested style embedding models with no fine-tuning, to measure their off-the-shelf ability to identify whether two documents were written by the same author and report results using the standard ROC-AUC metric used in AV. In Table \ref{table:avevaltable}, we compare the performance of \textsc{StyleDistance} embeddings against LISA and the \citet{styleemb} style embeddings. On average, \textsc{StyleDistance} outperforms the other representations on AV, demonstrating its effectiveness in representing style.

\avevaltable

\paragraph{Automatic Style Transfer Evaluation} \citet{lowresourceauthorshipstyletransfer} proposed the LUAR embedding model as an automatic measure for ``style transfer accuracy''. This approach was effective and was subsequently adopted for style transfer evaluation \citep{astropop,paraguide,tinystyler}. However, the LUAR model considers both style and content, hence confounding two aspects of style transfer evaluation---accuracy and meaning preservation---which are typically measured separately. Content-independent style representations would be a better measure for this task. We use the same evaluation dataset of 675 task instances used by  \citet{lowresourceauthorshipstyletransfer} where given an example text of a target author's style, the task is to discriminate which of two texts (a style transfer output and another actual text by the target author) is written by the target author. We show our results on this task in Table \ref{table:styletransfereval}. \textsc{StyleDistance} proves to be a more effective discriminator than all models, including LUAR. All models surpass human performance in distinguishing style transfer outputs. Since automatic style transfer evaluation typically includes a separate score for meaning preservation, using a model like LUAR (which is not content-independent) for measuring style transfer accuracy undermines the rigor of the style transfer accuracy metric. We find that a robust content-independent model like \textsc{StyleDistance} may enhance automatic style transfer evaluation by: (1) acting as a stronger discriminator, and (2) ensuring style transfer accuracy is assessed independently of meaning preservation.

\paragraph{Style Transfer Steering} Previous systems, like TinyStyler, have leveraged style embeddings to steer style transfer \citep{tinystyler}. While the original TinyStyler system rewrites text by conditioning on \citet{styleemb} embeddings, we demonstrate that \textsc{StyleDistance} provides an alternative, and reproduce TinyStyler with \textsc{StyleDistance} embeddings. We showcase an example of an output in Table \ref{table:smalltinystyleroutputs} with more details and results in Appendix \ref{sec:appendix:tinystyler}. We will make this version of TinyStyler available as a resource. With this result, we demonstrate \textsc{StyleDistance} can be used as a simple drop-in replacement for downstream applications in systems where weaker style representations have been previously used.
\styletransferevaltable
\smalltinystyleroutputs

\section{Conclusion}
\label{sec:conclusion}

We introduced \textsc{StyleDistance}, a novel method for training content-independent style embeddings using synthetic parallel examples. By employing a large language model to generate a dataset of near-exact paraphrases with controlled style variations, we overcome limitations associated with content leakage and imperfect parallel examples in existing style embedding methods. Evaluations using the STEL and STEL-or-Content tasks demonstrate that embeddings trained solely on synthetic examples can capture style extremely well. Additionally, the technique's ability to generalize to unseen style features indicates its potential to represent a broader range of style attributes beyond those addressed in synthetic data generation. Notably, our results highlight the efficiency of large language models in creating task-specific representations. This approach circumvents the need for manual dataset collection and the reliance on weak implicit assumptions over data, offering a more direct and accurate method for training on the precise task-specific features of interest.

\paragraph{Model, Data, and Code} We release the \textsc{StyleDistance} models, the  \textsc{SynthStel} dataset, and our code for other researchers to use at: {\small\url{https://huggingface.co/StyleDistance/}}.
\section * {Limitations}

Our approach shows strong results using 40 style features across 7 categories, though it does not fully cover the near-infinite range of possible style variations. The synthetic data may introduce some systematic biases, and we observe occasional repetitive patterns in the generated examples. Nonetheless, our method outperforms existing style representations, and we find that training on our selected 40 features offers strong generalization to unseen styles. Expanding this feature set could further enhance performance. While generating perfectly parallel examples for all style features is challenging---particularly for certain style features that may have overlap with content \citep{tstsurvey}---our model effectively leverages synthetic data to improve content-independence. Additionally, our current focus on sentence-level generations leaves room for future work to explore varying text lengths and multi-sentence style variations, which could further strengthen our approach.
\section * {Ethical Considerations}

This work demonstrates the potential of synthetic data for enhancing style embeddings. However, it is important to recognize that the synthetic data generated by large language models may reflect and reinforce existing biases inherent in these models \citep{lisa}. While our approach shows significant promise, ongoing efforts should ensure that such synthetic datasets are evaluated for fairness and bias to promote more equitable outcomes.
\section*{Acknowledgements}

This research is supported in part by the Office of the Director of National Intelligence (ODNI), Intelligence Advanced Research Projects Activity (IARPA), via the HIATUS Program contract \#2022-22072200005. The views and conclusions contained herein are those of the authors and should not be interpreted as necessarily representing the official policies, either expressed or implied, of ODNI, IARPA, or the U.S. Government. The U.S. Government is authorized to reproduce and distribute reprints for governmental purposes notwithstanding any copyright annotation therein.

\bibliography{custom}

% Appendices
\appendix\onecolumn
\section{Style Features and Definitions}
\label{sec:appendix:stylefeatures}
We list all style features selected for our synthetic dataset below along with the positive and negative prompts (used for constructing a full prompt for generating positive and negative examples as shown in Appendix \ref{sec:genposandneg}) and definitions (used to help define the style feature to human annotators in the annotation interface in Appendix \ref{sec:appendix:humanannotation}).

\stylefeaturestable

\newpage
\section{Generation Prompts and Details}
\label{sec:appendix:generationdetails}

Below we detail the structure of our prompts and inference parameters used for synthetic data generation.

\subsection{Extracting Topics from C4}
\label{sec:extracttopic}

We select random sentences from C4 \citep{t5andc4}, and extract the fine-grained topic of the sentence using GPT-4 \citep{gpt4} and the zero-shot prompt shown below. We perform sampling with \texttt{temperature} = 1.0 and \texttt{top\_p} = 0.0.

\vspace{1em}
{\small
\begin{verbatim}
 What is the fine-grained topic of the following text: {sentence} Only return the topic.    
\end{verbatim}}
\vspace{1em}

\noindent The fine-grained topic is then used as part of the attributed prompt described in Section \ref{sec:genposandneg} to ensure diversity in the generations.

\subsection{Generating Positive and Negative Example Sentences for Each Style Feature}
\label{sec:genposandneg}

For each style feature, we generate positive and negative parallel examples using a zero-shot prompt and the attributed prompt (AttrPrompt) method \citep{attrprompt} with GPT-4 to create diverse and realistic synthetic examples. To generate many examples per style feature, we use a new unused topic extracted from C4 in the prompt each time, we randomly sample a new permutation of the attributes in the prompt, and we perform sampling with \texttt{temperature} = 1.0 and \texttt{top\_p} = 1.0. We demonstrate an example below for the ``Usage of Active Voice'' style feature.

\vspace{1em}
{\small
\begin{verbatim}
Generate a pair of active and passive sentences with the following attributes:
    1. Topic: {topic}
    2. Length: {sentence_length}
    3. Point of view: {point_of_view}
    4. Tense: {tense}
    5. Type of Sentence: {sentence_type}
    
Ensure that the generated sentences meet the following conditions:
    1. There is no extra information in one sentence that is not in the other. 
    2. The difference between the two sentences is subtle. 
    3. The two sentences have the same length.
    {special_conditions_for_style_feature}
Use Format:
    Active: [sentence]
    Passive: [sentence]
Your response should only consist of the two sentences, without quotation marks.
\end{verbatim}}
\vspace{1em}

\noindent For the exact prompts for each style feature, see the code in our supplementary materials for this work.
\newpage
\section{Human Annotation Details}
\label{sec:appendix:humanannotation}

We provide an example of a task instance in our annotation interface. Human annotators were asked to rate whether the style feature was present or not in the sentence, with the option to also select "Possibly" if the annotator was unsure (instructed to use sparingly). We provide annotators with a definition of each style feature as well.

\mturkinterfacefig

\noindent We used a population of graduate students taking a class on natural language processing as the annotators. Each task instance was annotated by 10 distinct human annotators. We assign a score of 0 to ``No'', 0.5 to ``Possibly'', and 1 to ``Yes''. We average the scores from all  10 annotators assigned to each task instance. We consider to have  agreement for a positive example if the average score is >=0.5, and   for a negative example if the average score is < 0.5.

We measure inter-annotator agreement using Krippendorf's Alpha \citep{krippendorffalpha} which indicates moderate agreement of 0.55.  As a more easily interpretable measure of agreement between annotators, for each task instance, we also find, on average, around 8 out of the 10 annotators annotated in agreement on whether a style feature was present or not in the text.
\newpage
\section{Training Details}
\label{sec:appendix:trainingdetails}

\trainingdetailstable

More exact training details can be found in the source code provided in the supplementary materials for this work.
\newpage
\section{Visualization of \textsc{StyleDistance} Embedding Space}
\label{sec:appendix:umap}

We compare the embedding space of \citet{styleemb} and \textsc{StyleDistance} on informal/formal texts from GYAFC\footnote{ Grammarly’s Yahoo Answers Formality Corpus which contains 110K informal/formal sentence pairs.} \citep{gyafc} in Figure~\ref{fig:umapcomparison} below.

\umapfig
\newpage
\section{STEL and STEL-or-Content Task Visualization}
\label{sec:appendix:stelfig}

\stelfig

\newpage
\section{STEL and STEL-or-Content Results on \textsc{SynthStel}}
\label{sec:appendix:fullsynthstelresults}

\fullsynthstelevaltable
\newpage
\section{Style Feature Ablation Details}
\label{sec:appendix:ablationdetails}

For the generalization experiment and ablation results we demonstrate in Table \ref{table:ablationeval}, we list the style features ablated (removed from the training data) for the Out-of-Domain and Out-of-Distribution conditions.
\\~\\
\noindent \textbf{Out-of-Domain}:
\begin{enumerate}
    \item Usage of Formal Tone
    \item Usage of Contractions
    \item Usage of Numerical Substitution
    \item Complex Sentence Structure
    \item Usage of Text Emojis
    \item Usage of Emojis
\end{enumerate}
~\\
\noindent \textbf{Out-of-Distribution}:
\begin{enumerate}
    \item Usage of Formal Tone
    \item Usage of Polite Tone
    \item Fluency in Sentence Construction
    \item Usage of Only Uppercase Letters
    \item Usage of Only Lowercase Letters
    \item Incorporation of Humor
    \item Usage of Sarcasm
    \item Usage of Contractions
    \item Usage of Numerical Substitution
    \item Usage of Numerical Digits
    \item Complex Sentence Structure
    \item Usage of Long Words
    \item Usage of Text Emojis
    \item Usage of Emojis
    \item Presence of Misspelled Words
\end{enumerate}
\newpage
\section{Style Transfer Performance}
\label{sec:appendix:tinystyler}

\noindent TinyStyler is a style transfer system that reconstructs texts by conditioning on style embeddings \citet{tinystyler}. The original system is trained on the Reddit Million User Dataset (MUD) \cite{khan2021deepmetriclearningapproach} with a \textbf{four step} procedure:
\begin{enumerate}
    \item A modified T5 model \cite{t5andc4} is trained to reconstruct texts from paraphrases and style embeddings (Wegmann).
    \item This unsupervised model is then used to generate style transfer outputs between multiple authors from the original corpus.
    \item The resulting synthetic data is then filtered using style embedding (Wegmann) distance and meaning preservation metrics.
    \item Finally, a model is then fine-tuned on the resulting filtered dataset.
\end{enumerate}
\noindent We reproduce the TinyStyler procedure with the exact dataset and hyperparameters in the original paper. We make only one modification: replacing Wegmann embeddings with \textsc{StyleDistance} in the generation and filtering steps. We include the formality transfer evaluation results in Table \ref{table:tstyler-formal}. In these automatic evaluations, the  \textsc{StyleDistance} conditioned model performs comparably. We additionally include examples comparing model output in Table \ref{table:tinystyler-outputs}.

\begin{table}[H]
\fontsize{8}{10}\selectfont
\centering

\begin{tabular}{lcccccc}
\toprule
Method & Acc ($\rightarrow \textit{F}$,$\rightarrow \textit{I}$) &  Sim ($\rightarrow \textit{F}$,$\rightarrow \textit{I}$) & Fluency ($\rightarrow \textit{F}$,$\rightarrow \textit{I}$) & Joint ($\rightarrow \textit{F}$,$\rightarrow \textit{I}$) & GPT-2  \\
\midrule
$\textsc{TStyler}_{\textsc{Wegmann}}$ & 0.94 (0.90, 0.98) & 0.82 (0.81, 0.82) & 0.77 (0.83, 0.72) & 0.78 (0.77, 0.80) & 112.5  \\
$\textsc{TStyler}_{\textsc{StyleDistance}}$ &  0.96 (0.94, 0.98) & 0.80 (0.80, 0.81) & 0.77 (0.82, 0.73) & 0.80 (0.79, 0.80) & 112.4 \\
\bottomrule
\end{tabular}

\caption{We reproduce the automatic formality transfer evaluation procedure from TinyStyler \cite{tinystyler} on the GYAFC dataset \cite{gyafc}. $\rightarrow F$ corresponds to formal transfer, and $\rightarrow I$ corresponds to informal transfer.
}
\label{table:tstyler-formal}
\end{table}

\tinystyleroutputs

\newpage
\section{Resources}
\label{sec:appendix:resources}

\noindent We provide links and citations to resources used in this paper which provide license information, documentation, and their intended use. Our usage follows the intended usage of all resources.

~\\ ~\\ \noindent We utilize the following models:
\begin{itemize}
    \small
    \item $\text{GPT-4}$ (full model, accessed June, 2024) \citep{gpt4}
    \item RoBERTa (\texttt{roberta-base}) \citep{roberta,bert,sentencetransformers}
    \item Learning Universal Authorship Representations (LUAR) Embedding model \citep{luar}
    \item Style embedding model from \citet{styleemb}
    \item LISA model \citep{lisa}
    \item CoLA model (\texttt{textattack/roberta-base-CoLA}) \citep{cola}
    \item \texttt{all-mpnet-base-v2} Sentence Similarity model \citep{sentencetransformers} ~ (\url{https://huggingface.co/sentence-transformers/all-mpnet-base-v2})
\end{itemize}

~\\ \noindent We utilize the following datasets and resources:
\begin{itemize}
    \small
    \item STEL dataset \citep{stel}
    \item Contrastive Authorship Verification dataset \citep{styleemb}
    \item C4 \citep{t5andc4}
    \item LIWC \citep{liwc}
    \item PAN 2011 - 2015 Authorship Verification Datasets \citep{pan11,pan13,pan14,pan15} - Curated at: \url{https://huggingface.co/datasets/swan07/authorship-verification}
    \item GYAFC \citep{gyafc}
\end{itemize}

~\\ \noindent We utilize the following software:
\begin{itemize}
    \small
    \item DataDreamer \citep{datadreamer}
    \item Transformers \citep{transformers}
    \item Datasets \citep{datasets}
    \item PEFT \citep{peft}
    \item Sentence-Transformers \citep{sentencetransformers}
    \item NLTK \citep{nltk}
\end{itemize}

~\\ \noindent We estimate the total compute budget and detail computing infrastructure used to run the computational experiments found in this paper below:
\begin{itemize}
    \small
    \item 8x NVIDIA RTX A6000 / 100GB RAM / 16x CPU -- 80 hours
\end{itemize}

\end{document}